*Original Article*

# Liver Infection Prediction Analysis using Machine Learning to Evaluate Analytical Performance in Neural Networks by Optimization Techniques

P. Deivendran[1], S. Selvakanmani[2], S. Jegadeesan[3], V. Vinoth Kumar[4]

[1]*Department of IT, Velammal Institute of Technology Panchetti, Chennai.*
[2]*Department of IT, R.M.K. Engineering College, Chennai.*
[3]*Department of IT, Velammal College of Engineering & Technology, Madurai.*
[4]*Department of IT, Velammal Institute of Technology, Panchetti, Chennai.*

[1]*Corresponding Author : deivendran1973p@gmail.com*

*Abstract* - Liver infection is a common disease, which poses a great threat to human health, but there is still able to identify an optimal technique that can be used on large-level screening. This paper deals with ML algorithms using different data sets and predictive analyses. Therefore, machine ML can be utilized in different diseases for integrating a piece of pattern for visualization. This paper deals with various machine learning algorithms on different liver illness datasets to evaluate the analytical performance using different types of parameters and optimization techniques. The selected classification algorithms analyze the difference in results and find out the most excellent categorization models for liver disease. Machine learning optimization is the procedure of modifying hyperparameters in arrange to employ one of the optimization approaches to minimise the cost function. To set the hyperparameter, include a number of Phosphotase, Direct Billirubin, Protiens, Albumin and Albumin Globulin. Since it describes the difference linking the predictable parameter's true importance and the model's prediction, it is crucial to minimise the cost function.

## 1. Introduction
The liver is the biggest inner limb in our human deceased; it is behind in ribcage on the upper right half of the guts. A large portion of the liver's mass resides in the body's right half. The liver is a critical part of the assimilation and handling of nourishment. Liver cells produce bile, a greenish liquid that guides the processing from the breakdown of medications. There are there a category of diseases like Hepatitis A, B, and C. Most a human is eating or drinking something that is infected by fecal. It might not have some symptoms goes missing by themselves within a period of 6 months without any long-term harm. Hepatitis B is caused by defenseless gender or captivating drugs with communal gratuitous. The last solitary is the approach from impure blood that gets keen on the body. However, nonalcoholic fatty liver disease is when too much fat has built up inside the liver. The additional fact can cause irritation and cell injury in our liver. Acute liver collapse may occur. When one has an extended of liver disease, the liver quits functioning within a very tiny period. Cirrhosis is the swelling of scars in our liver and extra scars replacing the well parts of the liver. Most of the survey is composed of medical ex-pediments in addition to the skin will be a suitable and the sum of the dataset to be used.

Even if liver tissue has been severely damaged, the early stages of liver disease are incredibly difficult to detect. According to several restorative professionals frequently overlook the analysis of the illness. Early detection is crucial and critical to protecting the patient because this could lead to improper medicine and treatment. The major goal of this learning is to enhance the correctness of result prediction and lower the cost of finding in medicinal engineering. As an effect, we classify patients with liver complaints by means of a variety of classification methods.

## 2. Literature Survey
Y Yugal Kuma & G. Sahoo published a document based on the unusual categorization technique that has been used in the northeast region of the dataset collected liver defects (2013) [10]. The outcome to facilitate the DT algorithm has improved, the accuracy is more than [1] (2017 Sontakke, Sumedh, et al.) 82.45% com-pare to further algorithms, and it gives an accuracy of 97.27%. P.M.Goel recommended a document based on two categorization techniques, NB and FT, and used the Waikato location for information and scrutiny of the dataset [20] (2020 V Gupta, et al.). Naive Bayes has bigger 76.6% accuracy in FT Tree has a maximum of 74.24% correctness and fulfilled that Naive Bayes[13] has improved the accuracy and also compared to other sets of algorithms[14] (Bendi Venkata Ramana et al.)[7]. A. Singh et al. [6] viewed different intelligent models and their applications, including single, ANN, Fuzzy logic, CBR and GA, data mining techniques, etc. Integration of ANN-CBR, logic AIS-ANN-FL-2020 [4] is also presented. The entire type of liver diseases like cirrhosis liver, liver cancer, liver fibrosis, and hepatitis liver, fatty liver is discussed [15]. S. Petrovic et al. [9] built up a CBR framework for producing dosage anticipated treatment of new blood tumor patients by catching oncologists' experience in treating past patients[25]. The proposed CBR framework utilizes an adjusted Dempster–Shafer theory to fuse measurement arranges proposed by the most comparative cases recovered [24] in the year 2021-V. E. Ekong et al. [2] represented a neuro-fuzzy-CBR-driven decision support system in identifying diseases due to depression by utilizing the solutions of past cases[27]. The proposed hybrid framework structure presents a similarity coordinating driven neuro-fuzzy[8] engineering that gives adaptability to doctors measuring the seriousness levels of side effects and manifestations' class [22]. This work also proposes CBR and neuro-fluffy mixture systems for solving real-life problems [11]. Heba Ayeldeen planned an article to guess the liver phase by decision tree[13]; Cario University discussed the technique in 2021. Using the dataset and outcome show that the resolution of the classifier correctness is 92.5% [3].Somaya Hashem obtainable a document for analysis of liver disease [28]. In this article, they used two algorithms like, SVM & DT, and the backpropagation was used in the UCI machine repository dataset during the period from 2019-2021[17]. The SVM algorithm has an accuracy of 70.42%, which is better than the backpropagation accuracy is 74.2% [14].

To get the right liver diagnosis, Sanjay Kumar, along with Sarthak Katyal[2022], created a categorization model using several data removal algorithms. In order to provide the parameters like precision, recall, and accuracy, five algorithms were applied to the dataset. Five distinct algorithms, including K-means, K-NN, Naive Bayes, and Random Forest, were used in this model. The outcome of this model revealed that of all the algorithms, the random forest had the highest level of accuracy.

In a representation, is based on there are three machine learning techniques, including DBSCAN, K-means, and similarity propagation, were proposed by Varun Vats, Lining Zhang, and colleagues [5]. The proportional performance and evaluation of the three methods indicated before were conducted based on the Silhouette coefficient.

For feature selection, L.Alice Auxilia[27] employed the Pearson coefficient. Decision Trees, Naive Bayes, Random Forest, SVM, and ANN, were utilized to predict liver disease. Last but not least, it was demonstrated that the decision tree outperformed alternative classification techniques

## 3. Proposed Model

Biomedical science is most important in the field of machine learning to identify liver disease. Throughout advances, several mechanisms of the learning algorithm are analyzed, where the computerized method to deal with the existing models deals with any kind of disease. This proposed method is used to find a number of machines that have entered our life. These approaches that can be used to find in the field of supervised and unsupervised education are two major methods for ML. The purpose of this method is to be used for data training and the other part for the test. The liver is the most interior limb of the human corpse. It was acting a major role for relocates blood throughout our stiff. The most matching cases from the previous cases can be retrieved by using KNN computation. Categorization of algorithms is extensively used in different medicinal applications, including linear discriminate analysis, diagonal linear discriminate analysis, SVM and K-NN algorithm.

Two variables are used; X and Y represent the two independent variables used to the extent level similar to the supposed and space ratio weighing machine. The above figure 1 is used for creating data processing logic using. In the below table-1, SVM is a supervised knowledge technique used for together categorization and weakening. The SVM classifier is to distinguish involving members of two classes in the preparation and testing of the dataset simultaneously. The major remoteness of the adjoining data point of generalization error will become less compared to the existing one.

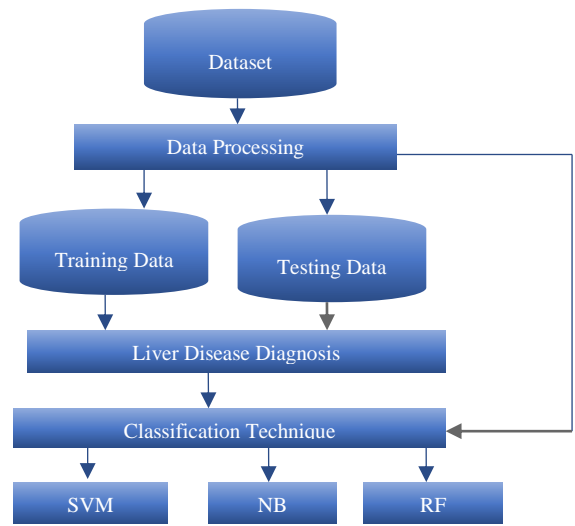

**Fig. 1 Functional Diagram of Liver Disease Classification**

**Table 1. Correlation Relations of Liver Disease Features**

| | | | | | | | | |
|---|---|---|---|---|---|---|---|---|
| Aspirate aminotransferase | 1 | 0.013 | 0.0083 | 0.09 | -0.092 | -0.04 | -0.19 | -0.28 | -0.21 |
| Billirubin | 0.013 | 1 | 0.93 | 0.23 | 0.24 | 0.26 | -0.0082 | -0.03 | -0.19 |
| Direct Billirubin | 0.0073 | 0.89 | 1 | 0.26 | 0.26 | -0.28 | -0.0017 | -0.26 | -0.18 |
| Phosphotase | 0.07 | 0.23 | 0.26 | 1 | 0.17 | 0.16 | -0.0024 | -0.24 | -0.22 |
| Aminotransferase | 0.783 | 0.23 | 0.26 | 0.15 | 1 | 0.16 | -0.085 | -0.063 | -0.068 |
| Direct aminotransferase | 0.03 | 0.24 | 0.28 | 0.19 | 0.75 | 1 | -0.089 | -0.079 | -0.071 |
| Protiens | 0.18 | -0.084 | -0.0015 | -0.029 | 0.24 | 0.64 | 1 | 0.81 | 0.23 |
| Albumin | 0.18 | -0.25 | -0.19 | -0.19 | -0.04 | -0.089 | 0.84 | 1 | 0.67 |
| Albumin Globulin | 0.28 | -0.12 | -0.23 | -0.23 | 0.0043 | -0.0065 | 0.23 | 0.73 | 1 |
| Types | Age | Billirubin | Direct Billirubin | Phosphotase | Amino-transferase | Direct Amino-transferase | Protiens | Albumin | Albumin Globulin |

Now the dataset points are measured in the type of {($y_1$, $x_1$), ($y_2$, $x_2$), ($y_3$, $x_3$)…… ($y_i$, $x_i$)} Here $y_i$=1/-1 is a stable which donates the group to which zi belongs, and i is the amount of samples dataset, where yi in the hyperplane of training data. K-NN categorization algorithm is single of the dataset analyzed in various parametric used. It is fast, and the preparation phase is minimal, but the testing phase is costly, and the cost is both in terms of time and storage. The dataset can be in the form of a multidimensional vector scaling form. The value can be measured by positive and negative classes and will be either + or -.

To find the value of $X_2 = X_1$-r[df/dx] at $X_1$. For all the points: $X_1$, $X_2$, $X_3$, $X_{i-1}$, $X_i$.

Table 1 above is used to determine the patient's age, albumin rate and Protiens. Billirubin range is vey from 0.013 to 0.28. However, the albumin ranges from 0.18 to 0.67. Similarly, the other parameter is albumin globulin in the low range from 0.12 to 1. So all possible ways to calculate by using Adam optimization techniques. The data size will be increased. All the data ranges from 0.18 to 0.0065. Finally, the dataset of the storage will be increased.

These outcomes will not necessarily be the same across all datasets. Despite some academics' claims to the contrary, SVM does not exhibit the highest accuracy in this situation. A researcher named C. Chuang [13] also offers an integrated paradigm. Outperformed every single and other integrated model, although, in this study, we found that integrated models do not always perform better than single and integrated models, as demonstrated in Tables 2 and 4. As a result, this study's findings show that an algorithm's success depends entirely on the dataset, type of data, amount of observations, dimensions, and decision boundary.

Age, sex, steroids, antiviral, fatigue, malaise, anorexia, large, firm liver, palpable spleen, spiders, ascites, varies, bilirubin, phosphate, albumin, protime, and histology are the independent variables. The dependent variable, on the other hand, represents a class with either category 1 (DIE) or category 2. (LIVE).

The above figure 2 shows the best parameter identification for different types of age groups using Adam optimizer techniques. Generally, all kinds of optimization techniques can be broadly classified into three categories like Exhaustive, Gradient and Genetic. The first one is used to locate the possible ways to get more solutions, but database size is small and has high accuracy. The second one is used to find the optimized data using the hyperparameter, even though the learning rate is very large, so skipping unwanted redundant data. The final one is used to find a good solution in the shortest time, but it is heuristics to get an optimal solution. Using the model, training can be classified into two categories big learning and small learning. Those models allowed us to update all kinds of parameters like Protiens, Direct Billirubin, Phosphotase and Albumin Globulin. All the parameters to evaluate on a training set of the dataset with treating optimization as a black box are used to minimize the functionality of the existing Adam Optimized.

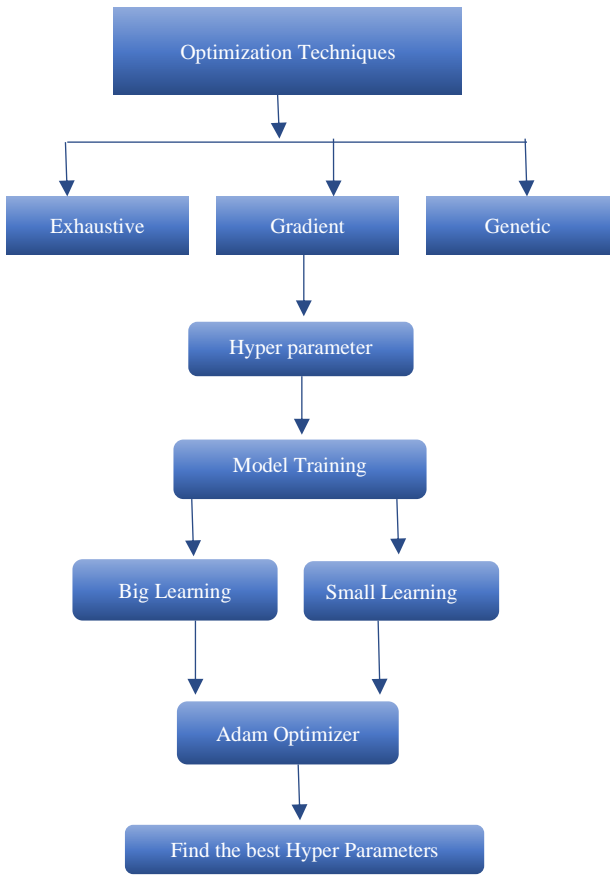

**Fig. 2 Best Parameters Identification Using Optimization Techniques**

**Patient:** positive in case of disease
**Healthy:** negative in case of disease

TN = the number of test cases effectively distinguished as fit.

FN = the number of cases erroneously distinguished as healthy.

TP = the number of cases effectively distinguished as the patient.

FP = the number of cases erroneously distinguished as patient

Accuracy=TP+TM/TP+FP+TN+FN

The precision of each test case of the capacity to separate the patient and sound accurate, to find the true positive and true negative in proportion foal all cases using the above mathematical function.

## 4. Result and Discussion

The specificity of analysis is its aptitude to decide the strong cases accurately. In the direction of computing, the amount of TP and strong cases are expressed by Sensitivity=TN/TN+FP.

The sensitivity is a set of capacities to decide the patient cases, and the proportion of the TN able to be expressed as Specificity=TP/TP+FN

The compassion of analysis is the capability to decide for the patient. It can be expressed by true positive and patient cases. In the case of Positive Prediction, the value is the proportion of positive results, and diagnostic test cases are true positive results, respectively, which can be represented as follows.

PPV=TP/TP+FP.

In the case of Negative prediction, the cost is the number of unenthusiastic results in diagnostics tests that are TN results, respectively, can be expressed by NPV=TN/TN+FN.

The above table 2 shows the different types of all algorithms for Liver harm using different datasets. The KNN achieved elevated accuracy of 100%, followed by SVM at 99.23% and LR at 68.62%, which is the subsequent best exactness. So the KNN is measured as the greatest algorithm for all the above mention datasets. The DT be achieved at the high accuracy of 68.62% of the classifies utilizing the dataset. The LR is a characteristic removal analysis, wherever the conversion means a set can be concentrated. In this paper, the evaluation method performs classification to measure each performance approach using Logistic Regression. KNN algorithm in the preparation stage just stored the dataset. At the same time, we get novel information and then classify that data into similar to the other dataset—the LR is also the best performer in terms of performance at 68.62%.SVM is an SVM algorithm capable of being second-hand for classification or failure problems.

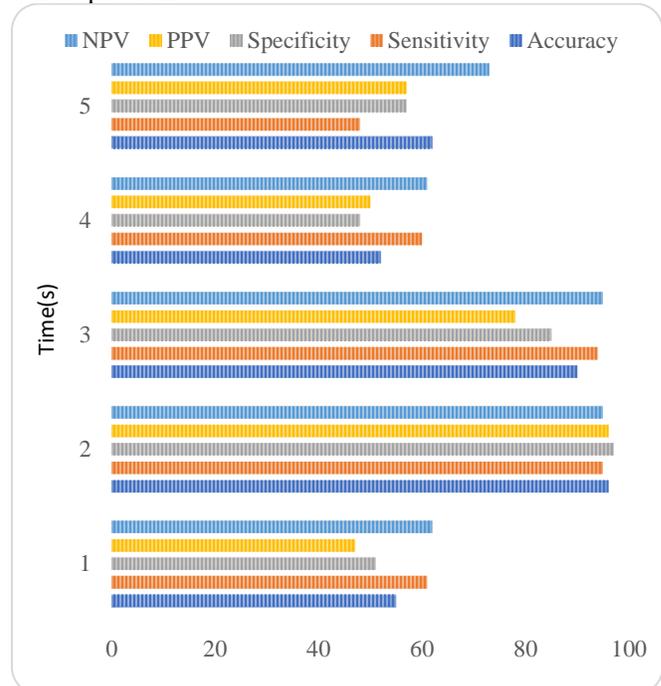

**Fig. 3 Existing Performance analysis**

Table. 2 Levels of Performance and Evaluation

| S.No. | Integrated | Model | LR | KNN | SVM | DT | RF |
|---|---|---|---|---|---|---|---|
| 1. | Accuracy (%) | Training | 57.32% | 100% | 92.52% | 54.32% | 64.52% |
| | | Testing | 58.27% | 98.3% | 92.61% | 56.67% | 76.23% |
| 2. | Sensitivity (%) | Training | 64.58% | 100% | 96.42% | 61.28% | 50.68% |
| | | Testing | 68.27% | 96% | 96.82% | 68.27% | 76.17% |
| 3. | Specificity (%) | Training | 54.72% | 100% | 89.34% | 51.72% | 60.32% |
| | | Testing | 53.58% | 94% | 90.51% | 50.48% | 62.28% |
| 4. | PPV (%) | Training | 51.32% | 100% | 78.37% | 52.32% | 61.12% |
| | | Testing | 50.61% | 92% | 72.51% | 54.51% | 63.51% |
| 5. | NPV (%) | Training | 67.46% | 100% | 97.62% | 64.16% | 75.26% |
| | | Testing | 68.62% | 90% | 99.23% | 66.32% | 75.42% |

Fig 3 shows the different types of algorithms to find the efficiency and accuracy using different parameters. It is used as a method in essential part to change our statistics and, subsequently, based upon these transformations to find the best possible limit. The various performance analyses of parameters like specificity, sensitivity, accuracy, positive and negative predictive values, and negative and positive predictive values are used. The PPV can increase to test the process reputedly in some situations. Say, for example, if we are affected by ELISA may be caused by some infection to test the population survey. The above graph shows that the accuracy ranges between 96% to 98%. If at all predictors the indication of the values, the accuracy of a normal and abnormal range ranges is 55% to 63%. The positive and negative values are calculated according to the patients and also divided by the number of patients with the true negative and true positive values of the parameter by means of accuracy. When we combine, the PPV of the test case is 73% and 96%, which is calculated as 1-(1-0.73)X(1-0.96).

The graph below represents the proposed performance model using different optimization techniques. Three parameters are used to find the performance of the patients according to the age groups Accuracy, Sensitivity and Specificity by using NPV and PPV algorithms. The minimum level of accuracy, Sensitivity, Specificity, NPV, and PPV is 65%, 51%, 60%, 61%, and 75. Similarly, the Average value of those 5 categories is 57%, 65%,55%,51% and 67%. The best value of the hyperparameter is the Maximum value of 99%,97%,99%,99% and 97. Different types of functionalities are used for best performance to search optimal value. Using bracketing algorithms, decent local algorithms, Gradient descent algorithms and second-order algorithms. Similarly, the non-differential objective functions are also possible to find the optimal algorithm like direct search algorithms and population algorithms for machine learning.

Most of the classification problems are used to find the optimal solution to prevent overfitting of the uncertainty and managed data. Still, now the computational process is difficult to solve the convex and non-convex regression techniques. The same way of handling the large-scale instances of the given value in SVM 68.62% has been discussed in the existing performance analysis. Applying the model training in the figure-2 for the performance analysis. A genetic algorithm to process data and selection in case of different types of age groups by applying the algorithm using the 8-binary number X1,0000000, Y1, 11111111, Z1, 10101010 and N1, 11110000. In-function optimization is used for the set of inputs to the given objective function, and the result has been either maximum or minimum value will be taken.

$$\text{The (parent of) } X1=\{0000|0000\}$$
$$(\text{Offspring-1})Y1=\{1111|1111\}$$
$$(\text{Offspring-2})Z1=\{1010|1010\}$$
$$(\text{Offspring-3}) N1=\{1111|0000\};$$

This input will satisfy the condition for the maximum number of iterations, temperature, objective, and acceptance in a given function. Set the number of iterations as follows using

Objective function of $F(x)=(X1, Y1, Z1……N1)^T$

**Genetic Algorithm and Iteration**
**Step-1:**
Binary value is as follows
    0 to 1 and 0 or 1 to 0.
    for routing where (X, Y) for all in N
**Step-2**
While (iteration)
    for loop over all the n dimensions nodes
    Generate (new) iteration
**Step-3**
Encode the input value
    Evaluate the new iteration
**Step-4**
end for
    process (Updated) iteration
    new(iteration)
    end while
**Step-5**
Decode and get the result.

Table 3. Comparison Result using PCA Dataset Existing System

| S.No. | Integrated | Model | LDA | DLDA | ADA | DQDA | SVM |
|---|---|---|---|---|---|---|---|
| 1. | Accuracy (%) | Training | 27.32 | 29.63 | 30.72 | 31.54 | 49.06 |
|  |  | Testing | 17.53 | 27.82 | 28.61 | 50.38 | **81.26** |
| 2. | Sensitivity (%) | Training | 46.21 | 49.41 | 65.84 | 71.50 | 49.32 |
|  |  | Testing | 51.43 | 48.31 | 65.35 | 49.64 | 94.71 |
| 3. | Specificity (%) | Training | 57.93 | 60.34 | 41.51 | 48.84 | 85.15 |
|  |  | Testing | 55.62 | 59.05 | 42.65 | 54.32 | 98.38 |
| 4. | PPV (%) | Training | 26.82 | 31.74 | 32.72 | 57.36 | 92.54 |
|  |  | Testing | 27.35 | 27.91 | 29.62 | 78.42 | 99.41 |
| 5. | NPV (%) | Training | 76.81 | 77.47 | 80.27 | 76.87 | 83.52 |
|  |  | Testing | 77.23 | 77.36 | 79.72 | 27.91 | 96.37 |

Table 4. Comparison Result using PCA Dataset Proposed System

| S.No. | Integrated | Model | LDA | DLDA | ADA | DQDA | KNN |
|---|---|---|---|---|---|---|---|
| 1. | Accuracy (%) | Training | 87.53 | 94.54 | 96.27 | 90.38 | 97.41 |
|  |  | Testing | 89.37 | 88.54 | 97.30 | 90.41 | **99.53** |
| 2. | Sensitivity (%) | Training | 78.37 | 92.74 | 99.19 | 91.42 | 99.13 |
|  |  | Testing | 77.38 | 89.67 | 98.41 | 92.69 | 98.48 |
| 3. | Specificity (%) | Training | 92.41 | 94.61 | 97.37 | 91.37 | 99.26 |
|  |  | Testing | 94.03 | 90.31 | 96.06 | 92.58 | 99.07 |
| 4. | PPV (%) | Training | 72.61 | 76.37 | 88.69 | 74.72 | 99.32 |
|  |  | Testing | 76.48 | 72.51 | 88.51 | 74.60 | 98.41 |
| 5. | NPV (%) | Training | 96.37 | 98.26 | 99.36 | 96.36 | 99.17 |
|  |  | Testing | 97.37 | 98.35 | 89.21 | 98.46 | 99.32 |

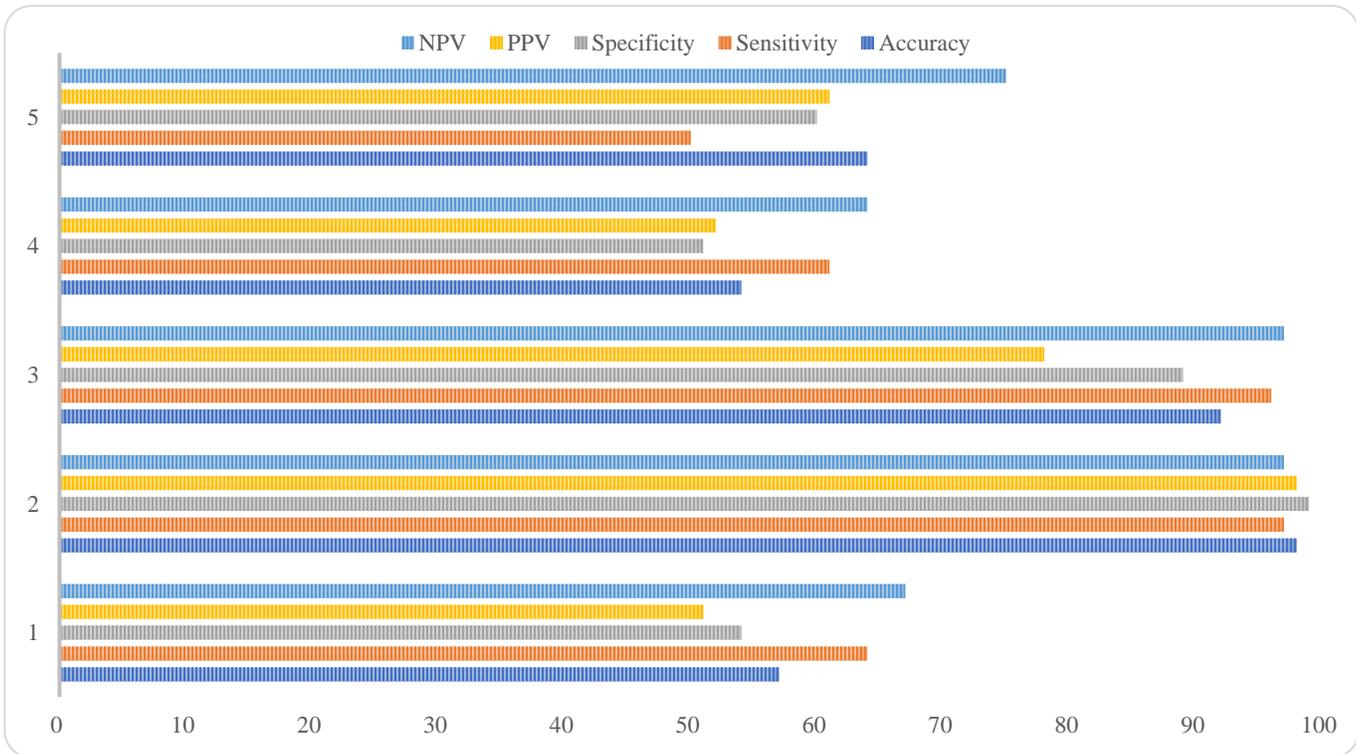

Fig. 4 Proposed Performance analysis

Suppose the random number is smaller than the bit changed or not changed to generate the string for the current value before swapping the digit. For example, some challenging values like tens, hundreds, thousands, or millions of input will pass on different objective functions. The problem of predictive function can be applied to ancient data to learn the predictions of the newly created data set. In a mapping function (X, Y) as the associated input value of any function Z=f(X, Y). Any given function of the inputs is weights to the different assumptions of the mapping function to solve the dataset in the number of variables passed to the best matches in the nearest neighbors of the prediction. However, many AI techniques are used to transform the data bits into text, images, audio or video based on the structure data.

## 5. Comparison Analysis of Svm and Knn Algorithm

In the below table-3 represent the five combinations of algorithms that were discussed. Using the dataset in LDa, the highest value of 77.23 % is the highest value of the model. Similarly, in a DQDA, the model value is 27.91 is the minimum value. The highest value is 78.48 in PPV aggregate value. For some classification techniques, the support vector machine has the highest value of 81.76 in an SVAM.

Table 4 is denoted by the Proposed system of the given dataset with a minimum value of 99.53. Similarly, in the subsequent value of the ADA, the highest value is 99.19, and the minimum value is 88.51. Suppose to take the minimum value is 98.36, the full record in a network system.

The Comparison analysis of Table.3 of the analysis is made, and the predicted value is determined by using the SVM algorithm; the result is 81.26%. However, the same analysis can be defined by using the KNN algorithm. The result has been improved to 99.53%.

## 6. Conclusion

This paper wills a useful analysis system for unusual sickness of liver diseases for patients to utilize. There are six characteristics of machine learning classifiers. Subsequently, all types of classifiers carry out about patient information by means of using various types of parameters, like LR, DT, SVM, LR, KNN, and RF classifiers. It gives the majority of high order accuracy 78%, reliant on F1 computing in the direction of calculating the liver sickness, and NP gives the smallest amount of precision, 57%. The performance of the classification technique will provide the decision support system. The relevance of the decision to calculate liver disease preceding and give an opinion of the health clause. This application be able to be profitable in small salaries to the nations where the absence of medical basics and just as meticulous specialist .